\documentclass[10pt, a4paper]{article}
\usepackage{lrec}
\usepackage{multibib}
\newcites{languageresource}{Language Resources}
\usepackage{graphicx}
\usepackage{tabularx}
\usepackage{soul}
\usepackage{booktabs}

\usepackage{epstopdf}
\usepackage[utf8]{inputenc}

\usepackage{hyperref}
\usepackage{xstring}
\usepackage{enumitem}

\usepackage{color}

\title{A Framework for Generating Annotated Social Media Corpora with Demographics, Stance, Civility, and Topicality\\ 
}

\name{Shubhanshu Mishra, Daniel Collier}

\address{shubhanshu.com, University of Illinois at Urbana-Champaign\\
        W.E. Upjohn Institute\\
         mishra@shubhanshu.com, Collier@upjohn.org\\
         }

\abstract{
In this paper we introduce a framework for annotating a social media text corpora for various categories. Since, social media data is generated via individuals, it is important to annotate the text for the individuals demographic attributes to enable a socio-technical analysis of the corpora. Furthermore, when analyzing a large data-set we can often annotate a small sample of data and then train a prediction model using this sample to annotate the full data for the relevant categories. We use a case study of a Facebook comment corpora on student loan discussion which was annotated for gender, military affiliation, age-group, political leaning, race, stance, topicalilty, neoliberlistic views and civility of the comment. We release three datasets of Facebook comments for further research at: \url{https://github.com/socialmediaie/StudentDebtFbComments}.\\ \newline }

\begin{document}

\maketitleabstract

 
 \section{Introduction}
 
\begin{figure*}[!h]
    \centering
    \includegraphics[width=\textwidth]{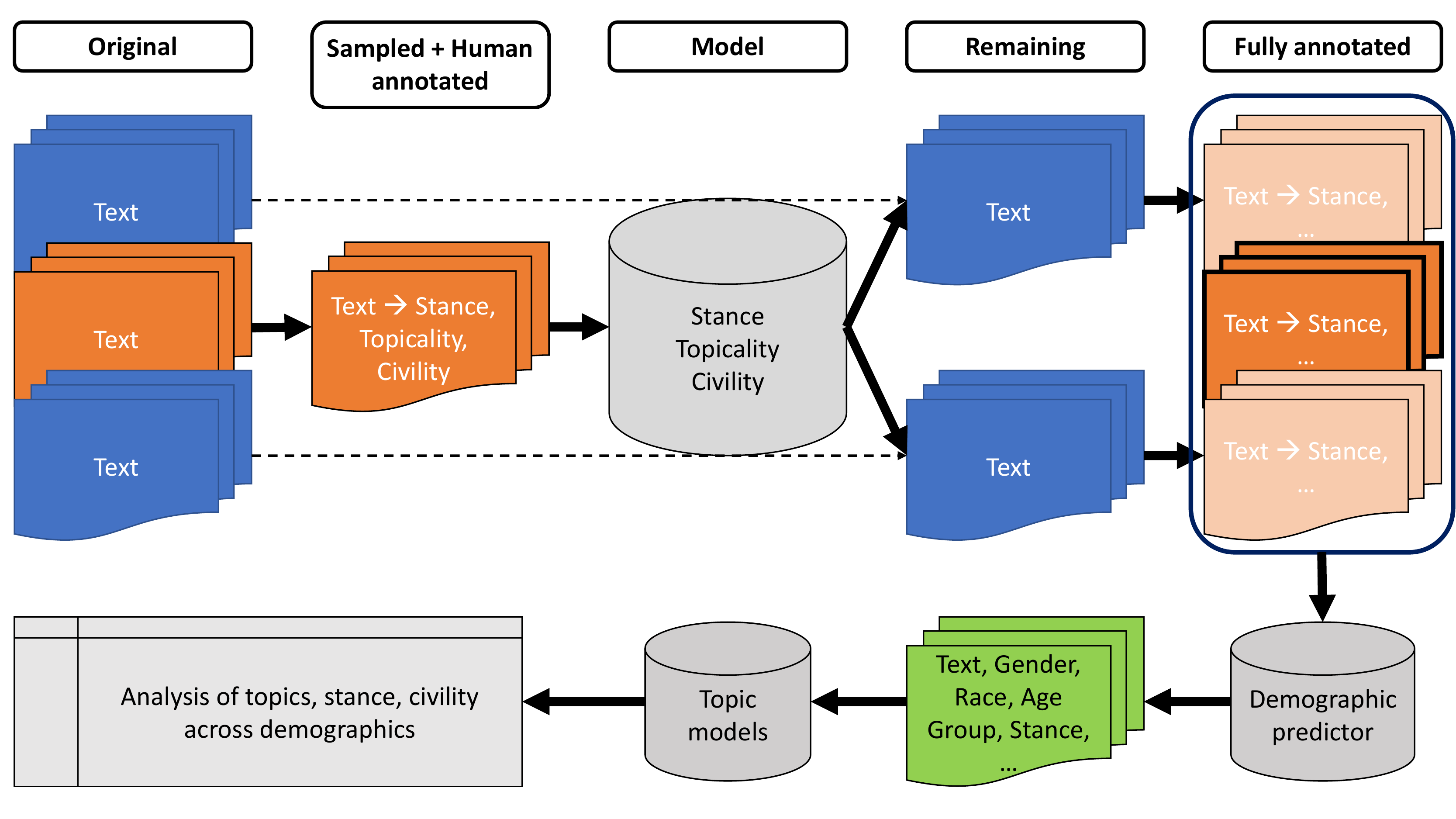}
    \caption{Workflow for data processing (Best viewed in color). Blue items are unannotated data, orange are human annotated data, light orange and data annotated using the model trained on annotated data.}
    \label{fig:workflow}
\end{figure*}

 Analysis of social media data can reveal interesting social patterns. These patterns can be utilized to study socio-technical systems and understand demographic differences in communication patterns \cite{Mishra2019a}. However, in order to make an inference which encompasses social attributes like race, gender, and age; as well as conversational attributes like stance, civility, and topicality; we need to identify these attributes in our corpus with high confidence \cite{Mishra2018SelfCite,Mishra2019HASOC}. This can be challenging if done manually. However, we can utilize machine learning techniques to iteratively add these attributes to all instances in our corpus. In this work we propose a framework (see Figure \ref{fig:workflow}) for enriching a social media corpora with social as well as conversational attributes by utilizing human annotations as well as machine learning techniques. 
 
 In this paper, we use three corpus of Facebook comments related to the topic of tuition free college. Our dataset is annotated for multiple categories ranging from demographic labels e.g. gender, race, military affiliation; to conversation based labels e.g. stance towards tuition free community colleges, neoliberalism \cite{Biebricher2012}, topicality, and civility. We release this human annotated corpora along with two larger corpus of Facebook comments related to the same topic but with labels predicted by models trained on the human annotated data\footnote{Workdone by SM while at UIUC}. 
 
\section{Workflow}

We defined a common workflow which can be applied to analyzing social media text by enriching it with user demographics and utilizing this enriched data to perform text based analysis. Our workflow as described in figure \ref{fig:workflow} is divided into stages which are described below in detail:

\subsection{Sub-sampling and human annotation}
In the sub-sampling and human annotation phase we select a small sample of the data. This can often be a $10\%$ random sample of the data, but if some attributes of the text data are known, e.g. source, demographics, then the sampling can be modified to ensure weighted sampling of the data. 

Next, the researcher should identify the list of categories the data should be annotated for. Importantly, these categories should be based on the research question which needs to be answered and should make sense given the type of the data. These categories should encompass labels which can be inferred from the text of with high confidence e.g. sentiment, stance, civility.

The human annotation process can often be carried out by multiple annotators and the annotation quality can be validated using various metrics such as \% agreement, kohen's $\kappa$, etc. Once, a high quality human annotated data is generated, we can move to the next stage. 

\subsection{Model based annotation of full data}
In this stage, the researcher should build a model of their annotations given the text. The goal here is to come up with a high accuracy and robust model of predicting the label from the text using the human-annotated data as a training set. There are various techniques for model building ranging from simple linear models on the n-grams in text to more complex neural network models. While the traditional training, development, and test split of the annotated data can be utilized to build these models, it can often result is a very small data being available for training the model. A more practical approach in this case is using k-fold cross validation to get the final model based on the full annotated data. Once, the model is available, we can safely annotate the full data using the models. 

\subsection{Demographic enrichment}
The demographic enrichment process utilizes the user information associated with the text to identify the demographic attributes of the author of the text. This can again be done using existing models which predict demographic attributes from the user's full name, inspection of user profile etc. The demographic identification can also happen in the human annotation phase where a human annotator inspects the user's public profile to identify their demographic attributes like gender, race, political leaning (based on pages liked or prior content posted on their profile), military affiliation, etc. 

\subsection{Analysis of text and labels}
In this final step, the researcher conducts their analysis to answer their research question, e.g. identifying the correlation between demographics and text categories such as stance, sentiment, etc.

Another analysis which can be done is topic modelling \cite{Blei2012} followed by topic correlations with demographics. 

\section{Case study: Identifying tuition free college debate via Facebook comments}

\begin{table*}[!h]
    \centering
\begin{tabular}{llrrr}
\toprule
       &             & sampled &   full &    new \\
\midrule
\textbf{Against/For} & \textbf{Against} &      2406 &  43309 &  25700 \\
       & \textbf{For} &      1437 &  20304 &   8011 \\
       & \textbf{Uncommitted} &        16 &      0 &      0 \\
\textbf{Age Category} & \textbf{29 and Under} &       369 &   3662 &   1486 \\
       & \textbf{30-49} &      1274 &  19614 &   9013 \\
       & \textbf{50 and Over} &      1532 &  31296 &  19441 \\
       & \textbf{Unknown} &       684 &   9041 &   3771 \\
\textbf{Civil/Uncivil} & \textbf{Civil} &      1875 &  29050 &  15095 \\
       & \textbf{Uncivil} &      1984 &  34563 &  18616 \\
\textbf{Gender} & \textbf{Female} &      1666 &  22248 &  10482 \\
       & \textbf{Male} &      2158 &  41365 &  23229 \\
       & \textbf{Unknown} &        35 &      0 &      0 \\
\textbf{Military Family} & \textbf{Military Family} &       759 &  11415 &   5647 \\
       & \textbf{Not Military Family} &      3093 &  52189 &  28063 \\
       & \textbf{Undetermined} &         7 &      9 &      1 \\
\textbf{Neoliberalism/Social Good} & \textbf{Neoliberalism} &      2402 &  42515 &  24610 \\
       & \textbf{Social Good} &      1315 &  19298 &   8463 \\
       & \textbf{Unknown} &       142 &   1800 &    638 \\
\textbf{OnTopic/Not-OnTopic} & \textbf{Not On-Topic} &      1455 &  24731 &  14280 \\
       & \textbf{On-Topic} &      2404 &  38882 &  19431 \\
\textbf{Political Leaning} & \textbf{Conservative Leaning} &      1892 &  39143 &  22926 \\
       & \textbf{Liberal Leaning} &       964 &  13185 &   5313 \\
       & \textbf{Undetermined} &      1003 &  11285 &   5472 \\
\textbf{Race } & \textbf{Asian} &        43 &    379 &    142 \\
       & \textbf{Black} &       332 &   5187 &   2293 \\
       & \textbf{International} &        79 &    831 &    269 \\
       & \textbf{Latino (a)} &       287 &   4042 &   1619 \\
       & \textbf{Middle Eastern} &        36 &    773 &    436 \\
       & \textbf{Unknown} &       126 &   1318 &    638 \\
       & \textbf{White} &      2956 &  51083 &  28314 \\
\textbf{Source} & \textbf{CNN} &       992 &  12832 &   7229 \\
       & \textbf{FOX} &       985 &  16665 &   7144 \\
       & \textbf{MSN} &       882 &  19312 &  13904 \\
       & \textbf{White House} &      1000 &  14804 &   5434 \\
\bottomrule
\end{tabular}
    \caption{Description of human-annotated data}
    \label{tab:data_desc}
\end{table*}

In this section we present a case study and its associated data-set some of which were utilized in our work \cite{Collier2019,Collier2019SSRN} to study the tuition free community college debate. We collected Facebook comments about tuition free community college topic after Barack Obama's America's College Promise (ACP) policy announcement. The data set was collected from Facebook comments made on posts related to this topic on the following Facebook's pages (referred as \textbf{Source}): 

\begin{enumerate}[noitemsep]
    \item CNN Money\footnote{Now called CNN Business \url{https://www.facebook.com/cnnbusiness/}}
    \item Fox's Sean Hannity\footnote{\url{https://www.facebook.com/SeanHannity/}}
    \item NBC News\footnote{\url{https://www.facebook.com/NBCNews/}}
    \item The White House\footnote{\url{https://www.facebook.com/WhiteHouse/}}
\end{enumerate}

These pages were selected as they attract a varying type of audience in terms of political alignment, and demographics.

We collected 66K comments (henceforth referred to as \textbf{full} dataset) between 8th and 16th January, 2015. From these comments the top 1000 comments (based on likes) per source were selected and these comments were further filtered for spam comments. This resulted in a dataset of 3859 comments (hence forth referred to as \textbf{sampled} dataset). We have also collected another dataset (henceforth referred to as \textbf{new} dataset) on this topic from comments posted on seven Facebook videos on popular news media pages regarding New York City's mayor Andrew Cuomo and Bernie Sanders announcement of a proposal for free tuition for state colleges in New York. The new dataset consists of 33K comments posted between September 2016 to February 2017 across all these videos. 

\subsection{Human annotation of sampled dataset}
In order to understand the social conversation around the tuition free college, we decided to annotate the sampled dataset with the following demographic and conversational attributes:

\begin{enumerate}[noitemsep]
    \item \textbf{Demographic Attributes:}
    \begin{enumerate}[noitemsep]
        \item Gender
        \item Race
        \item Political Leaning
        \item Military Family
        \item City and State
        \item Age Category
    \end{enumerate}
    \item \textbf{Conversational Attributes:}
    \begin{enumerate}[noitemsep]
        \item For/Against (Stance)
        \item Neoliberalism/Social Good (Stance)
        \item OnTopic/Not-OnTopic
        \item Uncivil/Civil
    \end{enumerate}
\end{enumerate}

The demographic attributes were selected based on their relevance to the issue of tuition free college. For example, people with military background or who are dependent of military veterans in US get various tuition waivers. Age category is also an important factor as the tuition policies and tuition fee have changed over time. Political leaning is also an important demographic attribute as it often governs a user's stance on a topic of social policy like tuition free college. The demographic attributes were identified via visiting the user's public profile and inferring the required information based on the content of the public profile. We have taken care to remove user identities when sharing this data publicly. Please see \cite{Collier2019SSRN,Collier2019} for more details about the labels. 

The conversational attributes focus on identifying user's stance on the topic e.g. if they are For or Against tuition free college. Similarly, comments were annotated for favoring Neoliberalism or social good policies. Next, comments were annotated if they were on the topic or off-topic by talking about some communal or unrelated policy to drive the conversation in a different direction. Finally, we also annotated if the comments used civil versus uncivil language.

A detailed breakdown of the various attributes in the sampled dataset are shown in table \ref{tab:data_desc}.

\subsection{Model based annotation of the full data}

\begin{table}[!h]
    \centering
    \begin{tabular}{lrr}
\toprule
{} &  overall &     mean (std. dev.) \\
\midrule
\textbf{Source                   } &     0.64 &  0.37 (0.05) \\
\textbf{Gender                   } &     0.74 &  0.59 (0.02) \\
\textbf{Age Category             } &     0.40 &  0.40 (0.00) \\
\textbf{Race                     } &     0.77 &  0.77 (0.00) \\
\textbf{Military Family          } &     0.80 &  0.80 (0.00) \\
\textbf{Political Leaning        } &     0.70 &  0.59 (0.01) \\
\textbf{Against/For              } &     0.95 &  0.84 (0.02) \\
\textbf{Neoliberalism/Social Good} &     0.90 &  0.76 (0.04) \\
\textbf{OnTopic/Not-OnTopic      } &     0.83 &  0.73 (0.02) \\
\textbf{Civil/Uncivil            } &     0.82 &  0.69 (0.03) \\
\bottomrule
\end{tabular}
    \caption{Micro F1 for various models. Overall means score on the full dataset after the best model has been identified using 5 fold cross validation. Mean is based on all CV splits for the model with best hyperparameters.}
    \label{tab:model_eval}
\end{table}

In order to conduct the analysis on the \textbf{full} dataset we first need to induce the attributes created in the sampled dataset to the full dataset. Since, the sampled dataset is from the same distribution as the full dataset we can assume this induction of attributes is likely to be accurate. To induce all the attributes, we train a logistic regression model using n-grams ($n \in \{1,2,3\}$) of the comments, and use this information to predict the attributes for the full dataset. The model is selected via 5 fold cross validation and the evaluation metric is micro F1 score. The mean evaluation score for the best model across all cross validation folds is reported in table \ref{tab:model_eval}. Furthermore, we also report an overall evaluation metric when the model is applied to the whole sampled data. From table \ref{tab:model_eval}, it is clear that there is a slight discrepancy between mean cross validation scores and overall score in some models. However, whenever the mean scores are high ($>0.7$), then the scores only vary by $0.1$. This means the models are still pretty accurate. We use the models trained on sampled dataset to annotate the comments from full and new dataset. A breakdown of the predicted attributes in the full and new dataset are shown in table \ref{tab:data_desc}.

\subsection{Demographic Enrichment}
\begin{table*}[!h]
    \centering
    \begin{tabular}{lrr}
\toprule
{Ethnicity/Gender} &   full &    new \\
\midrule
\textbf{Ethnicity=Uknown                                   } &    489 &    237 \\
\textbf{Asian-GreaterEastAsian-EastAsian     } &   1291 &    773 \\
\textbf{Asian-GreaterEastAsian-Japanese      } &    801 &    426 \\
\textbf{Asian-IndianSubContinent           } &   1605 &    778 \\
\textbf{GreaterAfrican-Africans             } &   1218 &    518 \\
\textbf{GreaterAfrican-Muslim              } &   1392 &    560 \\
\textbf{GreaterEuropean-British             } &  35733 &  18480 \\
\textbf{GreaterEuropean-EastEuropean        } &   1525 &    880 \\
\textbf{GreaterEuropean-Jewish              } &   7018 &   3635 \\
\textbf{GreaterEuropean-WestEuropean-French  } &   2538 &   1541 \\
\textbf{GreaterEuropean-WestEuropean-Germanic} &   1281 &    692 \\
\textbf{GreaterEuropean-WestEuropean-Hispanic} &   4605 &   2609 \\
\textbf{GreaterEuropean-WestEuropean-Italian } &   3056 &   2062 \\
\textbf{GreaterEuropean-WestEuropean-Nordic  } &   1061 &    520 \\
\midrule
\textbf{Gender=Male                                    } &  29627 &  14309 \\
\textbf{Gender=Female                                    } &  24423 &  14854 \\
\textbf{Gender=Unknown                                    } &   9563 &   4548 \\
\bottomrule
\end{tabular}
    \caption{Predicted gender and ethnicity}
    \label{tab:pred_gender_eth}
\end{table*}

Predicting race, and gender from the comment text is not the most appropriate as the signal to noise ratio is lower. However, user's full names provide a higher signal to predict their gender and race. In this step we utilize the US SSN gender database to predict the gender of the user \cite{Wais2016}. We use a 95\% confidence to assign a name as Male or Female based on the frequency with which the name occurs in the SSN database for a given gender label. In all other cases we assign a gender of Unknown. 

Ethnicity is identified using the name prism API \cite{Ambekar2009} \footnote{\url{http://www.name-prism.com/}}. The name prism API provides a hierarchical ethnicity prediction with three levels of predictions. 

The summary of gender and ethnicity predictions based on names for the full and the new dataset are shown in table \ref{tab:pred_gender_eth}. This denotes that our dataset has rich demographic diversity. 

More accurate methods of inferring gender and ethnicity can also be utilized as described in \cite{Mishra2018SelfCite}.

\subsection{Analysis of text and labels}

\newcommand*\rot{\rotatebox{90}}

\begin{table*}[!h]
    \centering
    \begin{tabular}{lrrrrrrrr}
\toprule
{} &     \rot{Against} &   \rot{For} &             \rot{Neoliberalism} & \rot{Social Good} &        \rot{Not On-Topic} & \rot{On-Topic} &         \rot{Civil} & \rot{Uncivil} \\
\midrule
\textbf{Ethnicity=Unknown                                   } &        0.67 &  0.33 &                      0.72 &        0.28 &                0.27 &     0.73 &          0.41 &    0.59 \\
\textbf{Asian-GreaterEastAsian-EastAsian     } &        0.72 &  0.28 &                      0.77 &        0.23 &                0.25 &     0.75 &          0.45 &    0.55 \\
\textbf{Asian-GreaterEastAsian-Japanese      } &        0.72 &  0.28 &                      0.75 &        0.25 &                0.24 &     0.76 &          0.47 &    0.53 \\
\textbf{Asian-IndianSubContinent            } &        0.69 &  0.31 &                      0.76 &        0.24 &                0.22 &     0.78 &          0.48 &    0.52 \\
\textbf{GreaterAfrican-Africans             } &        0.68 &  0.32 &                      0.74 &        0.26 &                0.25 &     0.75 &          0.45 &    0.55 \\
\textbf{GreaterAfrican-Muslim               } &        0.67 &  0.33 &                      0.73 &        0.27 &                0.25 &     0.75 &          0.45 &    0.55 \\
\textbf{GreaterEuropean-British             } &        0.77 &  0.23 &                      0.81 &        0.19 &                0.25 &     0.75 &          0.44 &    0.56 \\
\textbf{GreaterEuropean-EastEuropean        } &        0.78 &  0.22 &                      0.83 &        0.17 &                0.25 &     0.75 &          0.43 &    0.57 \\
\textbf{GreaterEuropean-Jewish              } &        0.78 &  0.22 &                      0.82 &        0.18 &                0.24 &     0.76 &          0.43 &    0.57 \\
\textbf{GreaterEuropean-WestEuropean-French  } &        0.74 &  0.26 &                      0.79 &        0.21 &                0.25 &     0.75 &          0.46 &    0.54 \\
\textbf{GreaterEuropean-WestEuropean-Germanic} &        0.77 &  0.23 &                      0.82 &        0.18 &                0.22 &     0.78 &          0.46 &    0.54 \\
\textbf{GreaterEuropean-WestEuropean-Hispanic} &        0.69 &  0.31 &                      0.75 &        0.25 &                0.23 &     0.77 &          0.43 &    0.57 \\
\textbf{GreaterEuropean-WestEuropean-Italian } &        0.76 &  0.24 &                      0.80 &        0.20 &                0.25 &     0.75 &          0.44 &    0.56 \\
\textbf{GreaterEuropean-WestEuropean-Nordic  } &        0.81 &  0.19 &                      0.84 &        0.16 &                0.23 &     0.77 &          0.44 &    0.56 \\
\midrule
\textbf{Gender=Female                                    } &        0.72 &  0.28 &                      0.77 &        0.23 &                0.21 &     0.79 &          0.49 &    0.51 \\
\textbf{Gender=Male                                    } &        0.80 &  0.20 &                      0.84 &        0.16 &                0.27 &     0.73 &          0.40 &    0.60 \\
\textbf{Gender=Unknown                                    } &        0.71 &  0.29 &                      0.76 &        0.24 &                0.24 &     0.76 &          0.45 &    0.55 \\
\bottomrule
\end{tabular}
    \caption{Distribution of conversational attributes across demographic attributes in \textbf{full} dataset}
    \label{tab:full_analysis}
\end{table*}

\begin{table*}[!ht]
    \centering
    \begin{tabular}{lrrrrrrrr}
\toprule
{} &     \rot{Against} &   \rot{For} &             \rot{Neoliberalism} & \rot{Social Good} &        \rot{Not On-Topic} & \rot{On-Topic} &         \rot{Civil} & \rot{Uncivil} \\
\midrule
\textbf{Ethnicity=Unknown                                  } &        0.70 &  0.30 &                      0.74 &        0.26 &                0.16 &     0.84 &          0.43 &    0.57 \\
\textbf{Asian-GreaterEastAsian-EastAsian     } &        0.74 &  0.26 &                      0.77 &        0.23 &                0.20 &     0.80 &          0.46 &    0.54 \\
\textbf{Asian-GreaterEastAsian-Japanese      } &        0.75 &  0.25 &                      0.77 &        0.23 &                0.22 &     0.78 &          0.45 &    0.55 \\
\textbf{Asian-IndianSubContinent            } &        0.71 &  0.29 &                      0.72 &        0.28 &                0.23 &     0.77 &          0.48 &    0.52 \\
\textbf{GreaterAfrican-Africans             } &        0.70 &  0.30 &                      0.75 &        0.25 &                0.21 &     0.79 &          0.48 &    0.52 \\
\textbf{GreaterAfrican-Muslim               } &        0.70 &  0.30 &                      0.74 &        0.26 &                0.24 &     0.76 &          0.46 &    0.54 \\
\textbf{GreaterEuropean-British             } &        0.74 &  0.26 &                      0.76 &        0.24 &                0.24 &     0.76 &          0.45 &    0.55 \\
\textbf{GreaterEuropean-EastEuropean        } &        0.75 &  0.25 &                      0.77 &        0.23 &                0.23 &     0.77 &          0.45 &    0.55 \\
\textbf{GreaterEuropean-Jewish              } &        0.73 &  0.27 &                      0.76 &        0.24 &                0.24 &     0.76 &          0.44 &    0.56 \\
\textbf{GreaterEuropean-WestEuropean-French  } &        0.74 &  0.26 &                      0.75 &        0.25 &                0.26 &     0.74 &          0.44 &    0.56 \\
\textbf{GreaterEuropean-WestEuropean-Germanic} &        0.72 &  0.28 &                      0.75 &        0.25 &                0.27 &     0.73 &          0.46 &    0.54 \\
\textbf{GreaterEuropean-WestEuropean-Hispanic} &        0.71 &  0.29 &                      0.74 &        0.26 &                0.23 &     0.77 &          0.44 &    0.56 \\
\textbf{GreaterEuropean-WestEuropean-Italian } &        0.73 &  0.27 &                      0.76 &        0.24 &                0.22 &     0.78 &          0.46 &    0.54 \\
\textbf{GreaterEuropean-WestEuropean-Nordic  } &        0.76 &  0.24 &                      0.77 &        0.23 &                0.26 &     0.74 &          0.40 &    0.60 \\
\midrule
\textbf{Gender=Female                                    } &        0.73 &  0.27 &                      0.75 &        0.25 &                0.23 &     0.77 &          0.46 &    0.54 \\
\textbf{Gender=Male                                    } &        0.75 &  0.25 &                      0.77 &        0.23 &                0.24 &     0.76 &          0.44 &    0.56 \\
\textbf{Gender=Unknown                                    } &        0.73 &  0.27 &                      0.74 &        0.26 &                0.22 &     0.78 &          0.44 &    0.56 \\
\bottomrule
\end{tabular}
    \caption{Distribution of conversational attributes across demographic attributes in \textbf{new} dataset}
    \label{tab:new_analysis}
\end{table*}

Once our dataset is enriched with all the attributes we can conduct different analysis to study the patterns of demographic variation of conversational attributes. In table \ref{tab:full_analysis} we present how the conversational attributes vary across demographics in the full dataset. It appears that users with male sounding names are $\sim 10\%$ more likely to be against the policy, $\sim 7\%$ more likely to make neoliberalistic arguments, $\sim 20\%$ more likely to use uncivil language, and $\sim 6\%$ more likely to comment not on the topic, compared to users with female sounding names. The gender analysis aligns with prior work on gender differences in ideologies \cite{Norrander2008}.

In table \ref{tab:new_analysis}, we replicate this analysis on the new dataset and find that the gender patterns are weaker compared to the full dataset. This discrepancy is likely due to domain shift when using a model trained on sample of full data to predict comments in the new data.

\section{Conclusion}

In this work, we have presented a workflow for analyzing social media text data using a mixture of conversational and demographic attributes. We show the usage of this workflow by presenting a case study on conversation in Facebook comments about tuition free college. The case study introduces three datasets on this topic including one human annotated dataset which was annotated for multiple conversational and demographic attributes. We enrich this data with gender and ethnicity predicted using user's full names. The language resources created by us can help in investigating the application of multi-label prediction on social media comments along with insights into demographic patterns of social conversation. We open source our dataset and code at \url{https://github.com/socialmediaie/StudentDebtFbComments}.

\section{Bibliographical References}
\label{reference}

\bibliographystyle{lrec}
\bibliography{lrec2020W-xample-kc}


\end{document}